\documentclass[a4paper
]{ceurart}

\usepackage{listings}
\usepackage[separate-uncertainty=true, per-mode=symbol]{siunitx}
\usepackage{enumitem}
\usepackage{algpseudocode}

\usepackage[caption=false, font=normalsize, labelfont=sf, textfont=sf]{subfig}

\begin{document}

\copyrightyear{2023}
\copyrightclause{Copyright for this paper by its authors.
  Use permitted under Creative Commons License Attribution 4.0
  International (CC BY 4.0).}

\conference{Proceedings of the Work-in-Progress Papers at the 13th International Conference on Indoor Positioning and Indoor Navigation (IPIN-WiP 2023), September 25 - 28, 2023, Nuremberg, Germany}

\title{Indoor Positioning based on Active Radar Sensing
	and Passive Reflectors: Concepts \& Initial Results}

\author[1]{Pascal Schlachter}[
email=pascal.schlachter@iss.uni-stuttgart.de
]
\cormark[1]
\address[1]{Institute of Signal Processing and System Theory, University of Stuttgart, Pfaffenwaldring 47, 70569 Stuttgart, Germany}

\author[2]{Zhibin Yu}[%
email=zhibinyu@huawei.com
]
\address[2]{Munich Research Center, Huawei Technologies Duesseldorf GmbH, Riesstraße 25, 80992 Munich, Germany}

\author[2]{Naveed Iqbal}[%
email=naveed.iqbal@huawei.com
]
\author[2]{Xiaofeng Wu}[%
email=xiaofeng.wu1@huawei.com
]
\author[1]{Sven Hinderer}[%
email=sven.hinderer@iss.uni-stuttgart.de
]
\author[1]{Bin Yang}[%
email=bin.yang@iss.uni-stuttgart.de
]

\cortext[1]{Corresponding author.}

\begin{abstract}
  To navigate reliably in indoor environments, an industrial autonomous vehicle must know its position. However, current indoor vehicle positioning technologies either lack accuracy, usability or are too expensive. Thus, we propose a novel concept called local reference point assisted active radar positioning, which is able to overcome these drawbacks. It is based on distributing passive retroreflectors in the indoor environment such that each position of the vehicle can be identified by a unique reflection characteristic regarding the reflectors. To observe these characteristics, the autonomous vehicle is equipped with an active radar system. On one hand, this paper presents the basic idea and concept of our new approach towards indoor vehicle positioning and especially focuses on the crucial placement of the reflectors. On the other hand, it also provides a proof of concept by conducting a full system simulation including the placement of the local reference points, the radar-based distance estimation and the comparison of two different positioning methods. It successfully demonstrates the feasibility of our proposed approach.
\end{abstract}


\maketitle

\section{Introduction}
Introduced in 1953, automated guided vehicles (AGVs) have been used extensively in industrial environments during the past decades. Recently, they are being replaced by autonomous mobile robots (AMRs). In contrast to their predecessors, AMRs are able to independently move through their environment without relying on tracks, predefined paths or supervision by human operators. In this way, they can increase flexibility, diversify applications and simplify the way of life of human beings within a broad field of different branches \cite{AMR4, AMRs}. In order to reliably navigate towards their destination and to safely avoid collisions with known obstacles, walls or other AMRs, they must determine their position. Hence, a dependable and accurate indoor vehicle positioning system is required. Therefore, in this paper we propose local reference point (LRP) assisted active radar positioning. It is a novel concept towards indoor positioning which is specifically tailored to the application on autonomous vehicles.

\section{Related Work}
All common approaches towards indoor vehicle positioning can be classified into three categories, namely fingerprint-based, geometry-based or SLAM-based methods.

First, fingerprint-based methods \cite{fingerprinting_survey} try to identify each position in an indoor environment by a unique set of significant features (a so-called fingerprint) of a wireless transmission channel. Such methods can be built upon existing communication systems (e.g. Wi-Fi) by using features that are already available like the channel state information (CSI) or the received signal strength indicator (RSSI). However, a large dataset of fingerprints must be collected by taking measurements on a dense grid. Moreover, this dataset must be updated each time anything in the environment changes.

Second, geometry-based methods determine the distances and/or angles to multiple active reference points and use the result to perform positioning via trilateration or triangulation. Most commonly, this is either done using the Bluetooth Low Energy (BLE) standard \cite{BLE_original_paper, BLE_AoA} or the ultra-wideband (UWB) technology \cite{UWB}. While BLE-based systems rely on measurements of the RSSI and/or the angles of arrival, UWB systems mostly use time-based measurements. Nevertheless, also geometry-based methods suffer from drawbacks for the application on AMRs such as a lack of accuracy (BLE) or the need of costly hardware and challenging calibration (UWB).

Third, simultaneous localization and mapping (SLAM) \cite{SLAM1, Visual_SLAM} can be applied. Its idea is to incrementally build a map of the unknown environment and use it for positioning at the same time. Thereby, the vehicle only relies on its onboard sensors scanning the environment and does not need an external reference system. 
%The simultaneous map and position estimation can be done based on Kalman, particle or information filters. Alternatively, a smoothing approach using least-square error minimization techniques can be applied.
However, like fingerprint-based methods, also SLAM-based methods struggle with non-stationary and large environments. Moreover, they can only estimate the relative position w.r.t. the simultaneously learned map.

Finally, sensor fusion, meaning the combination of multiple different positioning methods, can be applied in order to increase the accuracy and robustness. Especially, an iterative combination of the positioning results with the observations of an inertial measurement unit (IMU) is considered promising \cite{Fusion_IMU, Fusion_IMU_WIFI}. Thereby, the IMU captures the odometry data of the vehicle, meaning its change of position over time. Nevertheless, an IMU is not suited as a standalone positioning technique because it can only provide a relative position and its results typically drift away from the ground truth over time.

In conclusion, there is a lack of low-cost and easy to use but still highly accurate indoor positioning techniques. This especially holds for industrial applications like AMRs, since they typically require highly accurate estimates of the absolute position while operating in a regularly changing environment.

\section{Basic Idea}
The basic idea of our novel approach towards indoor positioning is to equip the environment with multiple low-cost passive retroreflectors such that a radar system mounted on an AMR observes a unique reflection characteristic at each position. In general, this reflection characteristic can include the distances, angles and relative velocities between the vehicle and the reflectors as well as the radar cross sections (RCSs) and polarimetrics of the corresponding reflectors. In the remainder of this paper, the reflectors will be denoted as local reference points (LRPs) and we will refer to the observed reflection characteristics as the LRP fingerprints.

Accordingly, our method combines the ideas of geometry- and fingerprint-based methods to get the best of both worlds. By using a well-designed radar system, our method is not only able to provide robust and highly accurate distance measurements to the LRPs compared to RSSI based estimates but also only requires cheap and simple passive instead of active reference points. Moreover, in contrast to existing fingerprint-based methods, the LRP fingerprint of each position can be calculated from the well-known positions and properties of the LRPs and does not change as long as the positions of the LRPs remain constant. In this way, our method does not need the extensive collection of a dataset and is also applicable in regularly changing environments.

In this initial work, we do not want to limit our approach to only work with high-end radar technologies. Instead, we propose and investigate a baseline which only requires a simple low-cost radar system with a single non-polarimetric TX and RX antenna, respectively. Showing that it can already provide reliable position estimates allows us to build onto this foundation and further develop our method in future work e.g. by additionally using angular or polarimetric information to further improve its accuracy.

In the following, we describe the four subtasks of the iterative positioning and navigation of an AMR in detail.

\subsection{LRP placement}
The first subtask is to place the LRPs into the environment. The resulting LRP layout is desired to provide a unique LRP fingerprint for each position. In this work, we restrict ourselves to the simplest setup, namely using only the set of distances to the LRPs as the LRP fingerprint for each position. By using a more advanced radar system, also other measurements can be incorporated into the LRP fingerprint.

Note that, in contrast to existing geometry-based indoor positioning methods or GNSS, the individual distances cannot be assigned to the respective LRP they belong to. Concretely, due to the application of passive instead of active reference points, the only way to distinguish between LRPs is to use different reflectors which differ in their RCSs. However, this is challenging in practice because of the large multi-path fading effects within indoor environments. Accordingly, we assume that at most two different types of LRPs can be distinguished. As a result, positioning via trilateration is not possible.

In the following section \ref{sec:4}, we analyze which constraints an LRP layout must fulfill to ensure a bijective mapping between the LRP fingerprints and the positions. Thereby, we must deal with a so far unsolved mathematical ambiguity problem. From all layouts which provide unique LRP fingerprints for all positions, we can then select the one which best fulfills secondary requirements like having large distances among the LRPs or keeping a distance to all walls.

\subsection{Radar transmission}
If the environment is equipped with LRPs, the iterative positioning and navigation of an AMR can start. The first step is the radar transmission. Although, we only assume a simple radar hardware which cannot capture angular or polarimetric information, measuring the relative velocity and RCS of targets besides their distances comes without additional cost. Both are helpful, for instance, to distinguish LRPs from other targets as described in more detail below. Moreover, the simple radar hardware does not limit the measurement accuracy. Instead, the resolution and maximal detectable range of the distance and relative velocity are defined by the choice of the radar system parameters. Thus, they must be designed carefully as they directly influence the achievable positioning accuracy.

We decide to apply a chirp sequence frequency modulated continuous wave (FMCW) radar system. It is state-of-the-art and, in contrast to other radar technologies, enables a joint estimation of the distances, relative velocities and RCSs of multiple targets at the same time. Obviously, this is required here as we intend to detect multiple LRPs simultaneously.

If multiple vehicles equipped with LRP assisted active radar positioning operate in the same environment, their respective radar systems interfere with each other. This can lead to ghost targets and complicates the detection of the LRPs. To mitigate the interference, well-known interference cancellation and coordination techniques like multiplexing, listen-before-talk or radar-specific approaches either based on classical signal processing \cite{mitigation_classical} or deep learning \cite{radar_dl} can be applied.

\subsection{Radar signal processing} \label{sec:signal_processing}
The radar signal processing estimates the distances, relative velocities and RCSs for all LRPs from the radar RX signal. Therefore, first the peaks within the range-Doppler spectrum are detected using a constant false alarm rate (CFAR) detector. Thereby, the position of each peak determines the distance and relative velocity of the corresponding target while the RCS is obtained from its magnitude. Subsequently, the detected targets must be classified to determine whether they correspond to an LRP or to other objects in the environment. Most easily, this can be done by comparing the observed RCSs to the well-known properties of the reflectors. Additionally, to increase robustness, we can track the distances of the targets over time by applying a motion model. Concretely, we use the relative velocities to calculate the changes of the distances assuming a straight movement with constant velocity. This assumption is sufficient if the time between two measurements is small compared to the velocity of the AMR. In this way, we can predict the distance to each target for the next iteration based on the current measurements. These predictions help to identify the LRPs within the target list. Moreover, by applying a Kalman filter, the predictions are combined with the actual distance measurements to smooth the results. In conclusion, tracking the targets over time increases the robustness against measurement, detection and classification errors. The resulting estimated LRP fingerprints are then used for positioning.

\subsection{Positioning}
In this work, we present two different approaches to find the position of the AMR based on the set of smoothed distance measurements to the LRPs (i.e. the estimated LRP fingerprint) output by the Kalman filter, namely using a look up table and AMCL. Both approaches are described in detail below.

\subsubsection{Look up table}
Most straightforwardly, the positioning can be achieved by using a look up table. Since the LRP layout is well-known, for each position the corresponding LRP fingerprint can be calculated easily. Now the inverse mapping is desired, meaning getting the position based on the fingerprint. Nevertheless, the latter can be obtained using the first mapping. Concretely, for each position the distances to the LRPs are calculated and the resulting fingerprints are stored in a look up table. Then, for an estimated distance set, we search for the corresponding entry in the look up table and finally get the position. However, note that it is not possible to store all infinitely precise positions into a look up table. Instead, the entries must be limited to a finite number. For instance, this can be achieved by quantizing the distances to a certain resolution. Then we get a finite number of grids which can be identified by their quantized LRP fingerprints. These fingerprints are then mapped to the center points of the corresponding grids to get a finite look up table. Thereby, obviously, the resolution of the quantization directly determines the size of the look up table. It must be chosen as a trade-off between the memory consumption and the positioning accuracy.

\subsubsection{Adaptive Monte Carlo localization}
\renewcommand{\figurename}{Alg.}
\begin{figure}[!t]
	\centering
	\begin{algorithmic}[1]
		\State Initialization: Sample $\underline{s}_0^i$, $i=1,\ldots, N_0$ from $p(\underline{z}_0)$ $\rightarrow$ $S_0$
		\For{$k=0,1,\ldots$}
		\State Weighting: $m_k^i=p(\hat{\mathcal{D}}_k|\underline{s}_k^i)$
		\State Weight normalization: $m_k^i=\frac{m_k^i}{\sum_{i=1}^{N}m_k^i}$
		\State Positioning: $\underline{\hat{z}}_k=\sum_{i=1}^{N}m_k^i\underline{s}_k^i$
		\State Resampling:
		\For{$n=1,\ldots, N_{k+1}$}
		\State Select sample from $S_k$ using pmf $p(\underline{s}_k^i)=m_k^i$ $\rightarrow$ $S_k'$
		\EndFor
		\State Propagation: $S_k'$ $\xrightarrow[\text{motion uncertainty}]{\text{odometry data}}$ $S_{k+1}$
		\EndFor
	\end{algorithmic}
	\caption{Algorithm of AMCL.}
	\label{alg:amcl}
\end{figure}

\renewcommand{\figurename}{Fig.}
\setcounter{figure}{0} 

While the look up table approach is straightforward, it is sensitive to erroneous or ambiguous LRP fingerprint estimates. Concretely, if one or multiple elements of the estimated LRP fingerprint are still inaccurate, the look up table directly leads to a wrong position estimate. To gain robustness, the idea is to use the odometry data of the AMR, meaning its movement information between two measurements, to additionally track the position over time. The odometry data can be either taken from the motion control commands which are passed to the drive unit of the AMR or measured by an IMU.

The position tracking is done by applying a particle filter based approach called Adaptive Monte Carlo Localization (AMCL) \cite{MCL, MCL1}. It recursively estimates the three-dimensional state vector $\underline{z}_k=[x_k, y_k, \theta_k]^T$, i.e. the position and orientation of the AMR at time instance $k$. Therefore, the posterior density $p(\underline{z}_k|X^k)$ of the current state conditioned on all previous LRP fingerprint estimates $X^k=\{\hat{\mathcal{D}}_j, j=1,\ldots,k\}$ is recursively approximated by a weighted set of $N_k$ random particles $S_k=\{\{\underline{s}_k^i, m_k^i\}, i=1,\ldots,N_k\}$ at each iteration $k$. Thereby, each particle $\underline{s}_k^i$ describes one possible state vector and its corresponding weight $m_k^i$ denotes the probability that this particle describes the true current state vector $\underline{z}_k$.

After drawing an initial population of $N_0$ particles from a prior distribution $p(\underline{z}_0)$, each particle is weighted according to the likelihood $m_k^i=p(\hat{\mathcal{D}}_k|\underline{s}_k^i)$. Hence, the weight of each particle $m_k^i$ denotes the likelihood that the LRP fingerprint estimate $\hat{\mathcal{D}}_k$ is observed given that the AMR is located at this respective particle $\underline{s}_k^i$. Finally, the weights are normalized such that they sum to one. Then they are used to do the positioning by calculating the weighted mean of the particles. Subsequently, we do a resampling by duplicating particles with large weights and discarding particles with small weights. This prevents a depletion of the population of particles since over time many particles drift far enough for their weight to become too small to contribute to the posterior. Furthermore, the number of particles $N_k$ can be adapted to the variance of the previous population of particles to achieve a good trade-off between accuracy and computational complexity. Finally, the movement of the AMR between two radar measurements is modeled by propagating the particles according to the odometry data. Thereby, we account for possible inaccuracies by adding a Gaussian distributed motion uncertainty.

The previously described propagation of the particles, update of the weights, position estimation and resampling is done in each iteration of the positioning and navigation process as shown in Algorithm \ref{alg:amcl}. Thereby, the population of particles more and more converges towards the true position of the AMR with each new iteration. After convergence, since the position estimate can only range within the set of particles, the positioning will still be sufficiently accurate even in the case of completely erroneous LRP fingerprint estimates. Moreover, in this way possible ambiguities, meaning different positions that observe the same LRP fingerprint, can be resolved. This enables the application of simpler LRP layouts.

\section{Placement of the Local Reference Points}
\label{sec:4}
\begin{figure*}[!t]
	\centering
	\subfloat[\label{fig:systematic_ambiguities_a}]{\includegraphics[width=0.35\linewidth]{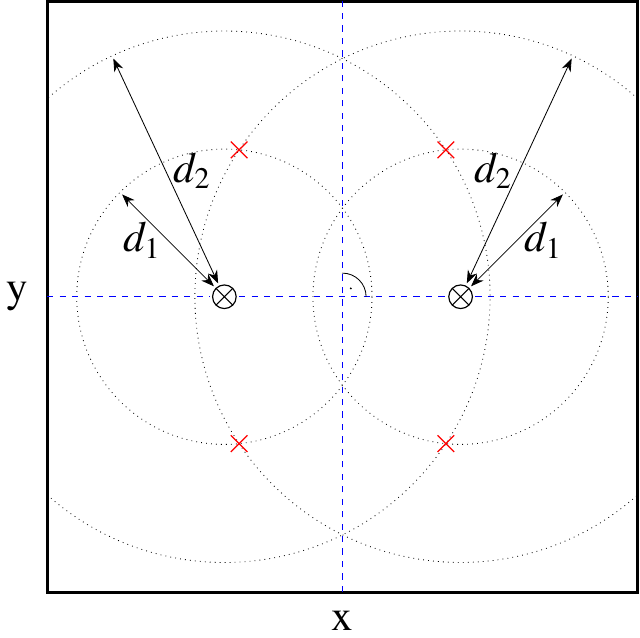}}\hspace{2cm}
	\subfloat[\label{fig:systematic_ambiguities_b}]{\includegraphics[width=0.35\linewidth]{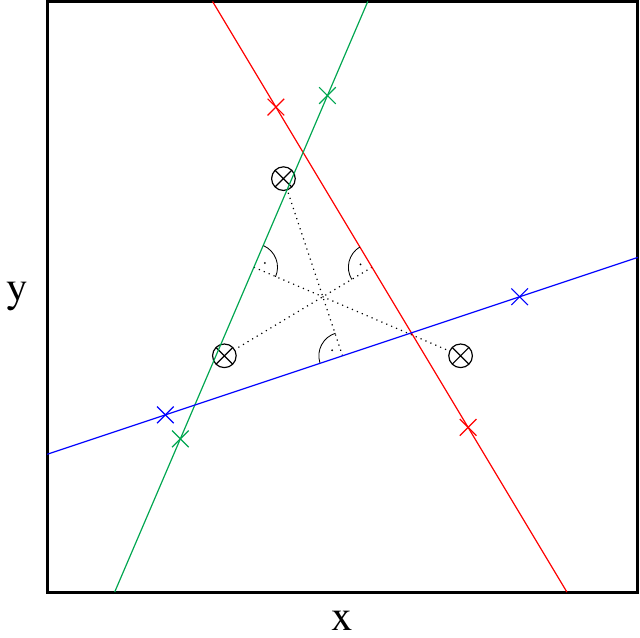}}\\
	\subfloat[\label{fig:systematic_ambiguities_c}]{\includegraphics[width=0.35\linewidth]{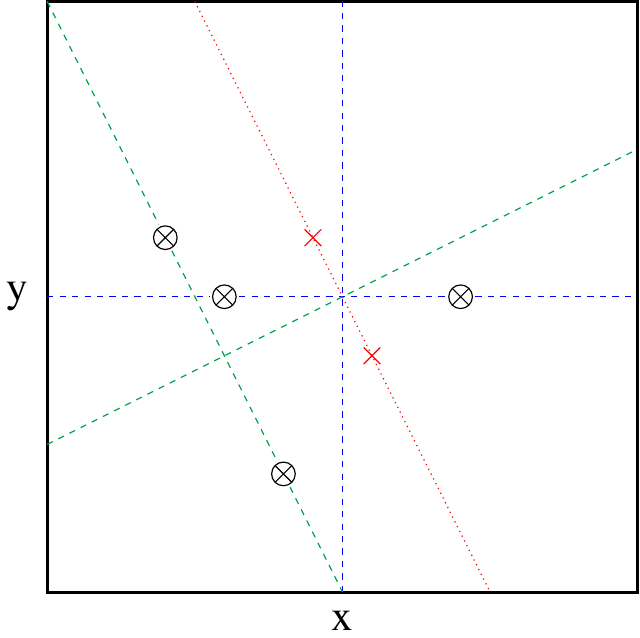}}\hspace{2cm}	\subfloat[\label{fig:systematic_ambiguities_d}]{\includegraphics[width=0.35\linewidth]{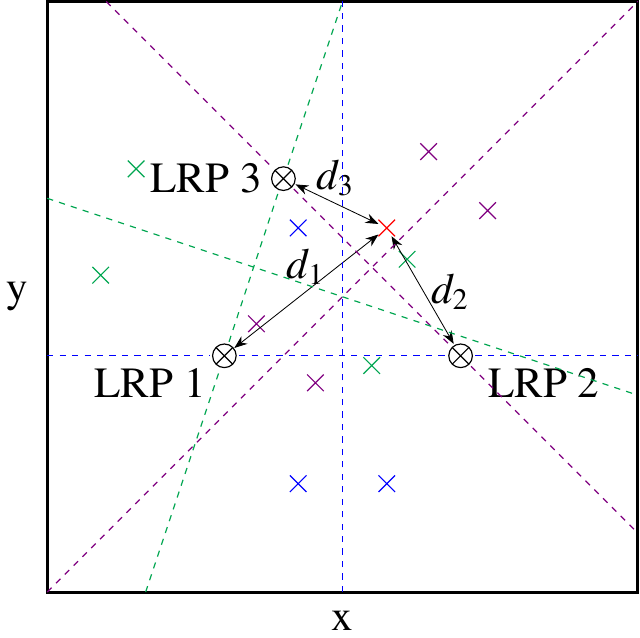}}
	
	\caption{(a)-(c) visualize the systematic ambiguities for exemplary LRP layouts using two, three and four LRPs of one type (black circles with crosses). Thereby, exemplary pairs of ambiguities are indicated by crosses of similar colors. 
		% (a) shows the symmetry axes (blue dotted lines) and a set having the same exemplary LRP fingerprint $\{d_1,d_2\}$ (red crosses). In (b), the three ambiguity lines (colored lines) and for each of them one arbitrary pair of ambiguous positions (colored crosses) are visualized. Finally, in (c), the ambiguity line (red dotted line) which results from a symmetry line of one combination of two LRPs crossing the symmetry point of the complementary combination is visualized.
		(d) supports the explanation of why random ambiguities occur by showing all positions whose LRP fingerprints have the distances $d_1$ and $d_2$ in common with the fingerprint of the arbitrary considered position (red cross).}
	\label{fig:systematic_ambiguities}
\end{figure*}
Now, we focus on one of the most crucial steps, namely the placement of the LRPs. Thereby, for a given indoor environment, the goal is to find a placement of $M$ LRPs
\begin{align}
	\mathcal{R}=\{(x_1, y_1, z_1, t_1)^T, \ldots, (x_M, y_M, z_M, t_M)^T\}
\end{align}
such that each position $p=(x, y)^T$ of an AMR can be characterized by a unique set of distances to the LRPs (LRP fingerprint) $\mathcal{D}$. In other words, the desired solution is an LRP layout $\mathcal{R}$ such that the mapping $\mathcal{D} \mapsto p$ is bijective. Thereby, $(x_n, y_n, z_n)$ denotes the position of the $n$-th LRP while $t_n$ describes its type. To the best of our knowledge this corresponds to a novel and so far unsolved mathematical ambiguity problem, since (in contrast to existing geometry-based methods or GNSS) we cannot uniquely assign the distances to their corresponding LRPs.

For our initial study in this work, we restrict ourselves to small-scaled rooms and make the following assumptions:
\begin{itemize}
	\item At most two types of LRPs can be distinguished reliably. Hence, $t_n\in\{0,1\}$, $n=1,\ldots,M$.
	\item Different LRPs of the same type cannot be distinguished.
	\item All LRPs are detectable from all positions of the room. This combines two assumptions. First, we do not consider shadowing, meaning that the direct view of the radar towards an LRP can be blocked. Second, we only consider environments where at each position all LRPs are within the maximal detectable distance of the radar system.
	\item All LRPs are mounted at the same height, meaning $z_1=\ldots=z_M=z$. Therefore it can be neglected and the problem can be reduced to the $xy$-plane.
\end{itemize}

Under these assumptions, we systematically investigated the following research questions (RQs) \footnote{A further analysis of these RQs under relaxed assumptions is done in \cite{sven} based on the results of this paper.}:
\begin{enumerate}[wide=10pt, leftmargin=\parindent, label=\textbf{RQ\arabic*}]
	\item Does an LRP layout exist which provides unique LRP fingerprints for all positions?
	\item If yes: How can we systematically find such a layout, i.e. which constraints must it fulfill and how many LRPs are necessary at least?
\end{enumerate}
Concretely, we started with one LRP and analyzed step by step how each additional LRP can help to avoid ambiguities, meaning different AMR positions with the same LRP fingerprint. Thereby, we found that in general two types of ambiguities are present. We call them systematic and random ambiguities. In the following we summarize our results. For more insights and detailed proofs, we refer to \cite{master_thesis} due to the limited space in this paper.

\subsection{Systematic ambiguities}
Systematic ambiguities denote all ambiguities which we can describe systematically. The foundation for all systematic ambiguities is the case of using two LRPs of one type. Thereby, intuitively, for each position in the room three other positions exist which have the same LRP fingerprint. These ambiguous positions are symmetric according to two symmetry axes. The first one is the line through the two LRPs and the second one is orthogonal to the first axis passing through the middle between the LRPs. In this way, the ambiguities are also point symmetric to the intersection point of the two axes. The symmetry axes (blue dotted lines) and one arbitrary set of ambiguities having the same LRP fingerprint $\{d_1,d_2\}$ (red crosses) are visualized in Figure \ref{fig:systematic_ambiguities_a} for an exemplary LRP layout. We can now use this knowledge to look at three and four LRPs of the same type as three and six combinations of two different LRPs each, respectively.

% Three LRPs
Adding a third LRP of the same type cannot resolve all these ambiguities. Concretely, for each combination of two LRPs there is a line containing point symmetric ambiguities which also have the same distance to the respective third LRP. As shown in Figure \ref{fig:systematic_ambiguities_b} for an exemplary LRP layout, each of these ambiguity lines (red, green and blue lines) passes through the center point between two LRPs (i.e. the symmetry point) and is orthogonal to the line connecting this point to the respective third LRP.

% Four LRPs
Finally, using a fourth LRP can resolve all systematic ambiguities. This is the case if for all six combinations of two LRPs, the symmetry axes do not cross the symmetry point of the respective complementary combination. Otherwise, as shown in Figure \ref{fig:systematic_ambiguities_c}, we again get (at least) a line of point symmetric ambiguities (red dotted line) which passes the symmetry point and is orthogonal to the corresponding symmetry axis.

% Two LRP types
Introducing a second type of LRP does not change much. When using three LRPs, one distance can now be assigned uniquely to the one LRP being of a different type. Nevertheless, still the ambiguity line corresponding to the combination of the two LRPs of the same type is left. When using four LRPs where two of each are of the same type, still no ambiguities are present if the previously stated condition holds. However, now we only need to check this condition for the two combinations of two LRPs of the same type.

% Summary
In summary, a layout with three or less LRPs always contains systematic ambiguities while a layout with four LRPs does not contain any systematic ambiguities as long as for all combinations of two LRPs, the symmetry axes do not cross the symmetry point of the respective complementary combination.

\subsection{Random ambiguities}
For layouts with three or four LRPs of one type, we find that there are additional ambiguities which we cannot describe systematically. We denote them as random ambiguities. So far, when searching for systematic ambiguities, for each combination of two LRPs and each position we only considered the three ambiguous positions w.r.t. this exact combination of two LRPs and checked under which circumstances they are also ambiguous w.r.t. the remaining LRP(s). E.g. in Figure \ref{fig:systematic_ambiguities_d}, when considering the combination of the LRPs 1 and 2 and searching for ambiguities of the arbitrary position indicated by a red cross, we only considered the positions indicated by blue crosses. However, there are additional ambiguity candidates having the exact same two distances in common with the considered position but w.r.t. one of the other combinations of two LRPs. Accordingly, in Figure \ref{fig:systematic_ambiguities_d} the green and purple crosses show the positions whose LRP fingerprints also contain the two distances $d_1$ and $d_2$ but w.r.t. the combination of the LRPs 1 and 3 or 2 and 3, respectively. Since we cannot assign the distances to the LRPs when using only one LRP type, these candidates may also result in ambiguities.

In contrast to the systematic ambiguities, here introducing a second type of LRP makes a difference. When using four LRPs where two of each are of the same type, the only ambiguity can stem from common ambiguities of the two complementary combinations of two similar LRPs. This corresponds to the case we already studied when searching for systematic ambiguities.

As a result, there are no random ambiguities when using layouts with two different types of LRPs. By combining this result with our findings for the systematic ambiguities we can answer \textbf{RQ1} and \textbf{RQ2} as follows: A layout using four LRPs of two types provides unique LRP fingerprints for all positions if the symmetry axes of one combination of LRPs of the same type do not cross the symmetry point of the other combination and vice versa. A detailed proof can be found in \cite{master_thesis}.

\section{Proof of Concept}
\subsection{Setup}
\begin{figure}[!t]
	\centering
	\includegraphics[width=.375\linewidth]{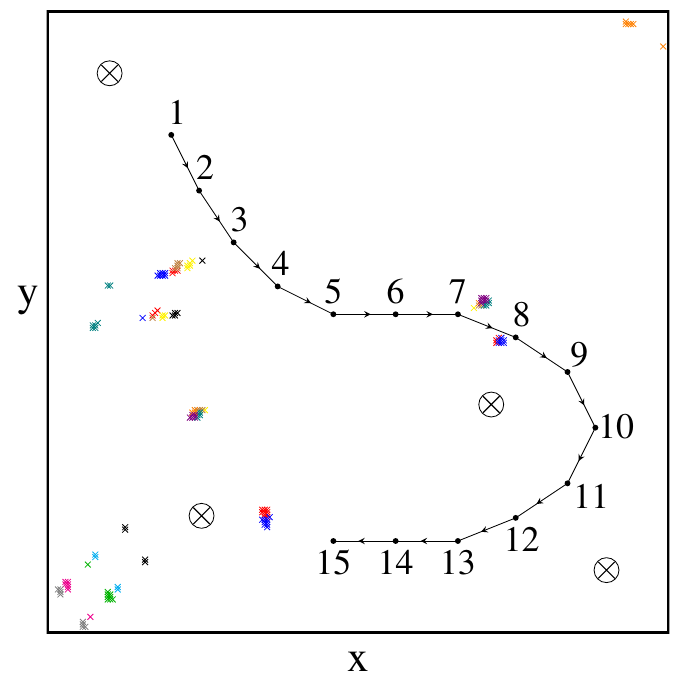}
	\caption{Arbitrary path of the AMR and selected LRP layout used for the system simulation. The colored areas indicate positions with the same LRP fingerprint. In the white areas, all positions have a unique LRP fingerprint.}
	\label{fig:layout}
\end{figure}

Finally, we conduct a system simulation to provide a proof of concept for our novel approach towards indoor positioning under idealized conditions of the environment and the LRPs. Concretely, we estimate the positions of an arbitrary path of an AMR within an empty room of size $\SI{5}{\meter}\times\SI{5}{\meter}\times\SI{4}{\meter}$. The path consists of 15 locations which are $\SI{0.5}{\meter}$ apart from each other, respectively. Furthermore, we assume the AMR to move linearly between two such locations with a constant velocity of $\SI{2}{\meter\per\second}$. Hence, the position is estimated every $\SI{0.25}{\second}$.

The room is equipped with four LRPs of one type which are all at the same height of $\SI{3}{\meter}$. We decide to use only one LRP type because distinguishing between different types of LRPs which only differ in their RCSs is challenging and error-prone in practice. Nonetheless, we desire a layout with as few ambiguities as possible. Accordingly, we generate $500$ random layouts, which fulfill the previously derived constraint to avoid systematic ambiguities and select the one with the smallest amount of random ambiguities. We expect the few remaining ambiguities to be resolvable by applying AMCL. The LRP layout and corresponding ambiguities together with the exemplary path of the AMR are shown in Figure \ref{fig:layout}.

The radar system parameters are designed such that it can measure distances up to $d_\mathrm{max}=\SI{19.125}{\meter}$ with a resolution of $\Delta d=\SI{0.075}{\meter}$ and relative velocities in the range $[\SI{-5.6816}{\meter\per\second},\SI{5.6816}{\meter\per\second})$ with a resolution of $\Delta v_r=\SI{0.3551}{\meter\per\second}$. The carrier frequency is selected to $f_c=\SI{60}{\giga\hertz}$. We simulate the radar transmission at the 15 exemplary positions by applying a ray tracer in a 3D model of the environment. Thereby, for the proof of concept, we neglect noise and diffused scattering effects while reflections and thus multipath fading are taken into account in the simulation. For each position, the ray tracer outputs the channel impulse response (CIR) and the Doppler shift for each channel tap, which we use to calculate the RX signal and finally the range-Doppler spectrum.

Subsequently, we find the peaks within the range-Doppler spectrum by applying a CFAR detector. Among the detected targets, we identify the four that correspond to the LRPs by a combination of comparing the estimated RCS of each target to the known value of the LRPs and tracking the targets over time. Concretely, the tracking is done by applying a linear motion model using the distance and relative velocity measurements of the previous position to predict the distances to the LRPs for the current position. The estimated LRP fingerprint then consists of the four most likely targets regarding both the RCS estimates and the distance predictions.

Finally, using the estimated LRP fingerprints, we compare both the look up table and the AMCL based position estimates with the true positions. Thereby, we choose the quantization resolution of the look up table entries equally to the distance resolution of the radar system, namely to $\SI{0.075}{\meter}$, resulting in $6574$ entries. Each estimated LRP fingerprint is then mapped to the entry with the smallest Euclidean distance. Regarding AMCL, we initially start with $N_0=10000$ uniform distributed particles. Concerning the propagation of the particles, we add Gaussian distributed motion uncertainties to both the turned angle and traveled distance contained in the odometry data. They have zero mean and the standard deviations $\sigma_\theta=\SI{5}{\degree}$ and $\sigma_d=\SI{5}{\centi\meter}$, respectively. For weighting the particles, $p(\hat{\mathcal{D}}_k|\underline{s}_k^i)$ is modeled as a Gaussian likelihood whose mean is the calculated LRP fingerprint corresponding to the respective particle $\underline{s}_k^i$. Moreover, its covariance matrix is chosen to $\boldsymbol{C}=(\SI{0.075}{\meter})^2\boldsymbol{I}$, where $\boldsymbol{I}$ denotes the identity matrix. Finally, to estimate the position of the AMR, we calculate the weighted mean of the $20$ best particles that are within a range of $\SI{15}{\centi\meter}$ to the particle with the largest weight.

\begin{figure*}[!t]
	\centering
	\includegraphics[width=\linewidth, trim={0 1.0cm 0 0.5cm},clip]{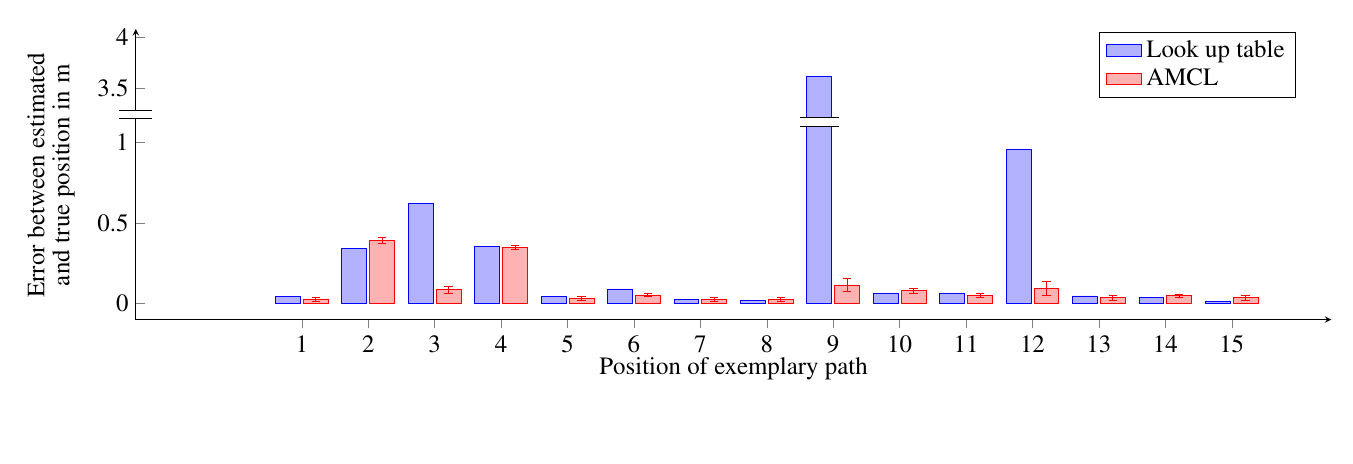}
	\caption{Errors between the estimated and true positions for all samples of the exemplary path using the look up table and AMCL, respectively. Thereby, for AMCL the mean and standard deviation of ten runs is reported. In contrast, the look up table approach is deterministic.}
	\label{fig:errors}
\end{figure*}

\subsection{Results} \label{sec:results}
The errors between the position estimates and the true positions for both the look up table approach and AMCL are shown in Figure \ref{fig:errors}. Thereby, the look up table provides deterministic position estimates while the results of the AMCL are subject to random deviations. Therefore, AMCL is performed ten times and we report the mean and standard deviation for each position.

The results prove two points. First and foremost, the system simulation successfully demonstrates the feasibility of our new approach towards indoor vehicle positioning and therefore provides a proof of concept. When using AMCL for positioning, the average of the mean errors for the 15 exemplary positions is at about $\SI{0.097}{\meter}$ only marginally larger than the range resolution of the radar system. Hence, our novel method proves to be able to reliably provide accurate position estimates.

Second, as expected, we observe AMCL to be significantly more robust than the look up table. As long as the estimated LRP fingerprints are correct and unambiguous, both positioning approaches achieve comparable accuracies. However, if they are erroneous or ambiguous the look up table approach fails completely while AMCL can still provide accurate position estimates. Concretely, the errors for the ninth and twelfth position using the look up table are around $\SI{3.5}{\meter}$ and $\SI{1}{\meter}$, respectively, while AMCL leads to errors of only around $\SI{0.1}{\meter}$ for both positions. Nevertheless, the full truth also includes the fact that this finding only holds once the whole set of particles $S_k$ has converged tightly enough towards the true position of the AMR. This typically takes a few iterations (here around six) which explains why for the positions two and four both positioning techniques perform equally poorly. Accordingly, in practice we recommend to initialize $S_0$ already as accurate as possible or to perform some ``warm-up'' iterations.

\section{Discussion about practical aspects}
\subsection{Robustness against multipath fading}
In indoor environments, the effect of multipath fading is generally stronger than outdoors since they are closed and narrow. While fingerprinting takes advantage of this, it is a challenge for most other radio transmission based indoor positioning methods in practice. Also the radar transmission of our proposed approach needs to cope with it. Accordingly, in the following, we will discuss the robustness of our method against multipath fading.

In terms of our application, multipath fading describes the effect that there are not only the desired direct paths from the radar TX antenna to the reflectors and back to the RX antenna. Instead, there are also countless other disturbing paths from the TX to the RX antenna like reflections from objects, the walls, the ceiling or even multi-reflections. First, we take a closer look at the direct paths. Their received power strictly follows the radar equation given as
\begin{align}\label{eq:radar}
	P_R=P_T\cdot\frac{G_TG_R\lambda_c^2}{(4\pi)^3r^4}\cdot\sigma~,
\end{align}
where $P_T$ and $P_R$ are the transmitted and received powers, respectively, $G_T$ and $G_R$ denote the antenna gains of the TX and RX antennas and $\lambda_c$ is the carrier wavelength of the radar system. Moreover, $r$ describes the distance between the radar and the reflector and $\sigma$ denotes the RCS of the reflector. Thereby, note that $P_T$, $G_T$, $G_R$ and $\lambda_c$ are all constant parameters of the radar system. Other single-reflection paths also follow this equation but of course the RCS $\sigma$ then is defined by the corresponding reflection surface/object while $r$ describes the distance to it. Multi-reflection paths, however, do not follow equation \ref{eq:radar}. Instead, they obviously need to consider multiple RCSs, one for each reflection, and also differ in their path loss which depends on the length of the path. Accordingly, the corresponding received power of a path experiencing $N$ reflections is calculated as
\begin{align}
	P_R=P_T\cdot\frac{G_TG_R\lambda_c^2}{(4\pi)^2d_0^2}\cdot\frac{\sigma_1}{4\pi d_1^2}\cdot\frac{\sigma_2}{4\pi d_2^2}\cdots\frac{\sigma_N}{4\pi d_N^2}~,
\end{align}
where $d_0$ is the length of the path between the TX antenna and the first reflection, $d_1,\ldots,d_{N-1}$ denote the lengths of the paths between the individual reflections and $d_N$ is the distance from the final reflection to the RX antenna. Furthermore, $\sigma_1,\ldots,\sigma_N$ are the RCSs corresponding to the $N$ reflections.

In order for our method to be robust against multipath fading, we face two challenges. First, the received powers of the desired direct LRP-paths must stand out enough compared to other paths with a similar travel time and to the noise floor such that they are detected as targets by the CFAR detector. Second, among all the paths that are detected as targets we must find the ones corresponding to the LRPs.

The keys to the robustness against multipath fading of our proposed radar-based approach, solving these two challenges, lie in the RCS of the reflectors and the RCS based LRP identification. Using appropriate and carefully designed reflectors with the largest possible RCS solves the first challenge. In this case, we expect the corresponding direct LRP-paths to be much stronger than reflections of random surfaces/objects and even more compared to multi-reflections. Accordingly, the following section \ref{sec:reflector} provides a discussion about the reflector choice. A further remedy can be to focus the radar transmission to a cone facing upwards. In this way, the number of disturbing paths is reduced by, among others, eliminating the possibly strong horizontal single-reflection paths. The second challenge is solved by the RCS based LRP identification together with the Kalman filter based LRP tracking as already introduced in section \ref{sec:signal_processing}. Since an FMCW radar measures the distance $r$ of each target based on the frequency shift, their RCSs can be calculated from the observed received power $P_R$ of the corresponding path according to equation \ref{eq:radar}. These estimations can then be compared to the well known RCS of the reflectors to identify the targets corresponding to the LRPs. Even in the unlikely cases that either another path which does not correspond to an LRP has the same estimated RCS as the reflectors or LRP targets are not detected by the CFAR detector, the Kalman filter based LRP tracking will help to still identify/recover the desired LRP targets.

The previously presented proof of concept simulation already showed that in this way our method is indeed robust against multipath fading. Thereby, the applied ray tracer simulated all paths between an omni-directional TX and the RX antenna within the simulation environment. Even in this worst-case-scenario, meaning a completely unfocused radar transmission in a narrow environment, the targets corresponding to the LRPs could be successfully detected and identified for nearly all positions. This can be seen from the good positioning accuracies for most positions using the look up table shown in Figure \ref{fig:errors}, since an erroneous LRP detection or identification directly leads to large positioning errors when using the look up table. Moreover, as already discussed in the previous section \ref{sec:results}, the simulations also showed that AMCL can still provide accurate position estimates for wrong LRP fingerprint estimates and in this way even further improves the robustness against the effects of multipath fading.

\subsection{Reflector choice} \label{sec:reflector}
In order to allow a reliable detection of the LRPs in a wide area, the retroreflectors are desired to have a large RCS and incidence angle. A first intuitive choice would be a spherical reflector due to its unlimited incidence angle. Its RCS is given as
\begin{align}
	\sigma_\mathrm{spherical}=\pi\cdot\left(\frac{d_0}{2}\right)^2~,
\end{align}
where $d_0$ denotes its diameter \cite{rcs_spherical}. Accordingly, for a sufficient RCS large reflectors are required which is undesirable in practice (e.g. for $\sigma_\mathrm{spherical}=\SI{0}{dbm^2}$ already a diameter of $d_0=\SI{1.13}{\meter}$ is required).

Alternatively, trihedral corner reflectors are simple yet effective and therefore widely used. Their RCS is given as
\begin{align}
	\sigma_\mathrm{corner}=\frac{4\pi a^4}{3\lambda_c^2}~,
\end{align}
where $a$ is the edge length of the trihedral and $\lambda_c$ denotes the carrier bandwidth of the radar system. Consequently, already a small reflector can provide a large RCS (e.g. assuming $\lambda_c=\SI{5}{\milli\meter}$ like in the simulation setup, then for $\sigma_\mathrm{corner}=\SI{0}{dbm^2}$ only a edge length of $a=\SI{5}{\centi\meter}$ is required). Although their incidence angle is limited they work well for angles within almost $\pm\SI{40}{\degree}$ \cite{RCS_trihedral}. Furthermore, multiple such reflectors can be assembled to further increase the covered area. Nonetheless, \cite{sven} shows that our novel approach towards indoor positioning still works well with LRPs having limited incidence angles by further analyzing the LRP positioning problem under this constraint based on the results of this paper.

Besides trihedral corner reflectors, promising choices combining sufficiently large RCSs and incidence angles are the Luneburg lens and the Van-Atta array \cite{bird2004design}. In further studies, we will experimentally evaluate which of the introduced retroreflectors is the best choice for our application.

\section{Conclusion and Outlook}
In conclusion, we proposed a novel approach towards indoor vehicle positioning. Its basic idea is to identify each position by the unique reflection characteristic (the so-called LRP fingerprint) an active radar system on top of an autonomous vehicle observes regarding a predefined layout of passive LRPs. To avoid having to make limiting assumptions regarding the radar system and to ensure robustness for further developments, we focused in this work on the set of distances to the LRPs as LRP fingerprints and assumed that at most two types of LRPs can be distinguished. Since this restriction does not allow the use of trilateration, we studied how to find an LRP layout which provides a unique LRP fingerprint for each position. Thereby, we found that a layout using four LRPs does not provide any systematic ambiguities if for all combinations of two LRPs, the symmetry axes do not cross the symmetry point of the respective complementary combination. Moreover, we showed that random ambiguities can be prevented by using two types of LRPs instead of only one. Finally, we provided a successful proof of concept by conducting a system simulation which demonstrated the feasibility of our proposed approach.

Further studies will be conducted in different directions. The first is the mitigation of the assumption that all LRPs are visible from all positions. This especially includes the extension to larger rooms, the application of reflectors with limited incident angles and the investigation of shadowing (some of this is examined in \cite{sven}). The second is to investigate the potential of additionally using angular and/or polarimetric informations. Are their benefits worth the additional costs? Finally, a hardware implementation can be helpful to enable experiments in a real-world scenario.

%%
%% Define the bibliography file to be used
\bibliography{refs}

\begin{thebibliography}{19}
\expandafter\ifx\csname natexlab\endcsname\relax\def\natexlab#1{#1}\fi
\providecommand{\url}[1]{\texttt{#1}}
\providecommand{\href}[2]{#2}
\providecommand{\path}[1]{#1}
\providecommand{\DOIprefix}{doi:}
\providecommand{\ArXivprefix}{arXiv:}
\providecommand{\URLprefix}{URL: }
\providecommand{\Pubmedprefix}{pmid:}
\providecommand{\doi}[1]{\href{http://dx.doi.org/#1}{\path{#1}}}
\providecommand{\Pubmed}[1]{\href{pmid:#1}{\path{#1}}}
\providecommand{\bibinfo}[2]{#2}
\ifx\xfnm\relax \def\xfnm[#1]{\unskip,\space#1}\fi
%Type = Article
\bibitem[{Oyekanlu et~al.(2020)Oyekanlu, Smith, Thomas, Mulroy, Hitesh, Ramsey,
  Kuhn, Mcghinnis, Buonavita, Looper, Ng, Ng’oma, Liu, Mcbride, Shultz,
  Cerasi, and Sun}]{AMR4}
\bibinfo{author}{E.~A. Oyekanlu}, \bibinfo{author}{A.~C. Smith},
  \bibinfo{author}{W.~P. Thomas}, \bibinfo{author}{G.~Mulroy},
  \bibinfo{author}{D.~Hitesh}, \bibinfo{author}{M.~Ramsey},
  \bibinfo{author}{D.~J. Kuhn}, \bibinfo{author}{J.~D. Mcghinnis},
  \bibinfo{author}{S.~C. Buonavita}, \bibinfo{author}{N.~A. Looper},
  \bibinfo{author}{M.~Ng}, \bibinfo{author}{A.~Ng’oma},
  \bibinfo{author}{W.~Liu}, \bibinfo{author}{P.~G. Mcbride},
  \bibinfo{author}{M.~G. Shultz}, \bibinfo{author}{C.~Cerasi},
  \bibinfo{author}{D.~Sun},
\newblock \bibinfo{title}{A review of recent advances in automated guided
  vehicle technologies: Integration challenges and research areas for 5g-based
  smart manufacturing applications},
\newblock \bibinfo{journal}{IEEE Access} \bibinfo{volume}{8}
  (\bibinfo{year}{2020}) \bibinfo{pages}{202312--202353}.
  \DOIprefix\doi{10.1109/ACCESS.2020.3035729}.
%Type = Article
\bibitem[{Alatise and Hancke(2020)}]{AMRs}
\bibinfo{author}{M.~B. Alatise}, \bibinfo{author}{G.~P. Hancke},
\newblock \bibinfo{title}{A review on challenges of autonomous mobile robot and
  sensor fusion methods},
\newblock \bibinfo{journal}{IEEE Access} \bibinfo{volume}{8}
  (\bibinfo{year}{2020}) \bibinfo{pages}{39830--39846}.
  \DOIprefix\doi{10.1109/ACCESS.2020.2975643}.
%Type = Article
\bibitem[{Khalajmehrabadi et~al.(2017)Khalajmehrabadi, Gatsis, and
  Akopian}]{fingerprinting_survey}
\bibinfo{author}{A.~Khalajmehrabadi}, \bibinfo{author}{N.~Gatsis},
  \bibinfo{author}{D.~Akopian},
\newblock \bibinfo{title}{Modern wlan fingerprinting indoor positioning methods
  and deployment challenges},
\newblock \bibinfo{journal}{IEEE Communications Surveys \& Tutorials}
  \bibinfo{volume}{19} (\bibinfo{year}{2017}) \bibinfo{pages}{1974--2002}.
  \DOIprefix\doi{10.1109/COMST.2017.2671454}.
%Type = Inproceedings
\bibitem[{Wang et~al.(2013)Wang, Yang, Zhao, Liu, and
  Cuthbert}]{BLE_original_paper}
\bibinfo{author}{Y.~Wang}, \bibinfo{author}{X.~Yang},
  \bibinfo{author}{Y.~Zhao}, \bibinfo{author}{Y.~Liu},
  \bibinfo{author}{L.~Cuthbert},
\newblock \bibinfo{title}{Bluetooth positioning using rssi and triangulation
  methods},
\newblock in: \bibinfo{booktitle}{2013 IEEE 10th Consumer Communications and
  Networking Conference (CCNC)}, \bibinfo{year}{2013}, pp.
  \bibinfo{pages}{837--842}. \DOIprefix\doi{10.1109/CCNC.2013.6488558}.
%Type = Article
\bibitem[{Pau et~al.(2021)Pau, Arena, Gebremariam, and You}]{BLE_AoA}
\bibinfo{author}{G.~Pau}, \bibinfo{author}{F.~Arena}, \bibinfo{author}{Y.~E.
  Gebremariam}, \bibinfo{author}{I.~You},
\newblock \bibinfo{title}{Bluetooth 5.1: An analysis of direction finding
  capability for high-precision location services},
\newblock \bibinfo{journal}{Sensors} \bibinfo{volume}{21}
  (\bibinfo{year}{2021}). \URLprefix
  \url{https://www.mdpi.com/1424-8220/21/11/3589}.
  \DOIprefix\doi{10.3390/s21113589}.
%Type = Article
\bibitem[{Alarifi et~al.(2016)Alarifi, Al-Salman, Alsaleh, Alnafessah,
  Al-Hadhrami, Al-Ammar, and Al-Khalifa}]{UWB}
\bibinfo{author}{A.~Alarifi}, \bibinfo{author}{A.~Al-Salman},
  \bibinfo{author}{M.~Alsaleh}, \bibinfo{author}{A.~Alnafessah},
  \bibinfo{author}{S.~Al-Hadhrami}, \bibinfo{author}{M.~A. Al-Ammar},
  \bibinfo{author}{H.~S. Al-Khalifa},
\newblock \bibinfo{title}{Ultra wideband indoor positioning technologies:
  Analysis and recent advances},
\newblock \bibinfo{journal}{Sensors} \bibinfo{volume}{16}
  (\bibinfo{year}{2016}). \URLprefix
  \url{https://www.mdpi.com/1424-8220/16/5/707}.
  \DOIprefix\doi{10.3390/s16050707}.
%Type = Article
\bibitem[{Durrant-Whyte and Bailey(2006)}]{SLAM1}
\bibinfo{author}{H.~Durrant-Whyte}, \bibinfo{author}{T.~Bailey},
\newblock \bibinfo{title}{Simultaneous localization and mapping: part i},
\newblock \bibinfo{journal}{IEEE Robotics Automation Magazine}
  \bibinfo{volume}{13} (\bibinfo{year}{2006}) \bibinfo{pages}{99--110}.
  \DOIprefix\doi{10.1109/MRA.2006.1638022}.
%Type = Article
\bibitem[{Fuentes-Pacheco et~al.(2015)Fuentes-Pacheco, Ruiz-Ascencio, and
  Rend{\'o}n-Mancha}]{Visual_SLAM}
\bibinfo{author}{J.~Fuentes-Pacheco}, \bibinfo{author}{J.~Ruiz-Ascencio},
  \bibinfo{author}{J.~M. Rend{\'o}n-Mancha},
\newblock \bibinfo{title}{Visual simultaneous localization and mapping: a
  survey},
\newblock \bibinfo{journal}{Artificial intelligence review}
  \bibinfo{volume}{43} (\bibinfo{year}{2015}) \bibinfo{pages}{55--81}.
%Type = Article
\bibitem[{Feng et~al.(2020)Feng, Wang, He, Zhuang, and Xia}]{Fusion_IMU}
\bibinfo{author}{D.~Feng}, \bibinfo{author}{C.~Wang}, \bibinfo{author}{C.~He},
  \bibinfo{author}{Y.~Zhuang}, \bibinfo{author}{X.-G. Xia},
\newblock \bibinfo{title}{Kalman-filter-based integration of imu and uwb for
  high-accuracy indoor positioning and navigation},
\newblock \bibinfo{journal}{IEEE Internet of Things Journal}
  \bibinfo{volume}{7} (\bibinfo{year}{2020}) \bibinfo{pages}{3133--3146}.
  \DOIprefix\doi{10.1109/JIOT.2020.2965115}.
%Type = Article
\bibitem[{Poulose et~al.(2019)Poulose, Kim, and Han}]{Fusion_IMU_WIFI}
\bibinfo{author}{A.~Poulose}, \bibinfo{author}{J.~Kim}, \bibinfo{author}{D.~S.
  Han},
\newblock \bibinfo{title}{A sensor fusion framework for indoor localization
  using smartphone sensors and wi-fi rssi measurements},
\newblock \bibinfo{journal}{Applied Sciences} \bibinfo{volume}{9}
  (\bibinfo{year}{2019}). \URLprefix
  \url{https://www.mdpi.com/2076-3417/9/20/4379}.
  \DOIprefix\doi{10.3390/app9204379}.
%Type = Article
\bibitem[{Hakobyan and Yang(2019)}]{mitigation_classical}
\bibinfo{author}{G.~Hakobyan}, \bibinfo{author}{B.~Yang},
\newblock \bibinfo{title}{High-performance automotive radar: A review of signal
  processing algorithms and modulation schemes},
\newblock \bibinfo{journal}{IEEE Signal Processing Magazine}
  \bibinfo{volume}{36} (\bibinfo{year}{2019}) \bibinfo{pages}{32--44}.
  \DOIprefix\doi{10.1109/MSP.2019.2911722}.
%Type = Article
\bibitem[{Geng et~al.(2021)Geng, Yan, Zhang, and Zhu}]{radar_dl}
\bibinfo{author}{Z.~Geng}, \bibinfo{author}{H.~Yan},
  \bibinfo{author}{J.~Zhang}, \bibinfo{author}{D.~Zhu},
\newblock \bibinfo{title}{Deep-learning for radar: A survey},
\newblock \bibinfo{journal}{IEEE Access} \bibinfo{volume}{9}
  (\bibinfo{year}{2021}) \bibinfo{pages}{141800--141818}.
  \DOIprefix\doi{10.1109/ACCESS.2021.3119561}.
%Type = Inproceedings
\bibitem[{Dellaert et~al.(1999)Dellaert, Fox, Burgard, and Thrun}]{MCL}
\bibinfo{author}{F.~Dellaert}, \bibinfo{author}{D.~Fox},
  \bibinfo{author}{W.~Burgard}, \bibinfo{author}{S.~Thrun},
\newblock \bibinfo{title}{Monte carlo localization for mobile robots},
\newblock in: \bibinfo{booktitle}{Proceedings 1999 IEEE International
  Conference on Robotics and Automation (Cat. No.99CH36288C)},
  volume~\bibinfo{volume}{2}, \bibinfo{year}{1999}, pp.
  \bibinfo{pages}{1322--1328}. \URLprefix
  \url{https://ieeexplore.ieee.org/document/772544}.
  \DOIprefix\doi{10.1109/ROBOT.1999.772544}.
%Type = Techreport
\bibitem[{Rekleitis(2004)}]{MCL1}
\bibinfo{author}{I.~Rekleitis}, \bibinfo{title}{A Particle Filter Tutorial for
  Mobile Robot Localization}, \bibinfo{type}{Technical Report}
  \bibinfo{number}{TR-CIM-04-02}, Centre for Intelligent Machines, McGill
  University, \bibinfo{address}{Montreal, Qu\'ebec, Canada},
  \bibinfo{year}{2004}. \URLprefix
  \url{http://www.cim.mcgill.ca/~yiannis/particletutorial.pdf}.
%Type = Article
\bibitem[{Hinderer et~al.(2023)Hinderer, Schlachter, Yu, Wu, and Yang}]{sven}
\bibinfo{author}{S.~Hinderer}, \bibinfo{author}{P.~Schlachter},
  \bibinfo{author}{Z.~Yu}, \bibinfo{author}{X.~Wu}, \bibinfo{author}{B.~Yang},
\newblock \bibinfo{title}{Indoor positioning based on active radar sensing and
  passive reflectors: Reflector placement optimization}
  (\bibinfo{year}{2023}).
%Type = Masterthesis
\bibitem[{Schlachter(2021)}]{master_thesis}
\bibinfo{author}{P.~Schlachter}, \bibinfo{title}{{Indoor Positioning based on
  Active Radar Sensing and Passive Reference Points}}, Master's thesis,
  University of Stuttgart, \bibinfo{address}{Stuttgart, Germany},
  \bibinfo{year}{2021}.
%Type = Article
\bibitem[{Rajyalakshmi and Raju(2011)}]{rcs_spherical}
\bibinfo{author}{P.~Rajyalakshmi}, \bibinfo{author}{G.~Raju},
\newblock \bibinfo{title}{Characteristics of radar cross section with different
  objects},
\newblock \bibinfo{journal}{International Journal of Electronics and
  Communication Engineering} \bibinfo{volume}{4} (\bibinfo{year}{2011})
  \bibinfo{pages}{205--216}.
%Type = Inproceedings
\bibitem[{Li et~al.(2010)Li, Zhao, Yin, Zhang, and Shan}]{RCS_trihedral}
\bibinfo{author}{C.~Li}, \bibinfo{author}{J.~Zhao}, \bibinfo{author}{J.~Yin},
  \bibinfo{author}{G.~Zhang}, \bibinfo{author}{X.~Shan},
\newblock \bibinfo{title}{Analysis of rcs characteristic of dihedral corner and
  triangular trihedral corner reflectors},
\newblock in: \bibinfo{booktitle}{2010 5th International Conference on Computer
  Science \& Education}, \bibinfo{year}{2010}, pp. \bibinfo{pages}{40--43}.
  \DOIprefix\doi{10.1109/ICCSE.2010.5593647}.
%Type = Inproceedings
\bibitem[{Bird(2004)}]{bird2004design}
\bibinfo{author}{D.~Bird},
\newblock \bibinfo{title}{Design and manufacture of a low-profile radar
  retro-reflector},
\newblock in: \bibinfo{booktitle}{NATO RTO SCI Symposium on Sensors and Sensor
  Denial by Camouflage, Concealment and Deception (RTO-MP-SCI-145). Brussels,
  Belgium}, \bibinfo{year}{2004}.

\end{thebibliography}

\appendix

\end{document}